\pdfoutput=1

\documentclass[11pt, final]{article}
\usepackage{tabularx}
\usepackage{longtable}

\newtheorem{definition}{Definition}[section]

\usepackage{times}
\usepackage{latexsym}
\usepackage[T1]{fontenc}

\usepackage[utf8]{inputenc}

\usepackage{microtype}

\usepackage{inconsolata}

\usepackage{graphicx}
\usepackage{hyperref}
\usepackage{natbib}
\usepackage{xcolor}
\title{CAPRAG: A Large Language Model Solution for \\ Customer Service and Automatic Reporting using\\ Vector and Graph Retrieval-Augmented Generation}

\author{
  \textbf{Hamza Landolsi\textsuperscript{1,2}},
  \textbf{Kais Letaief\textsuperscript{1}},
  \textbf{Nizar Taghouti\textsuperscript{1}},
  \textbf{Ines Abdeljaoued-Tej\textsuperscript{2,3}}
\\
\\
{\small \textsuperscript{1}INETUM Tunisia, Immeuble Harbour, Avenue de la Bourse, Tunis 1053, Tunisia,} \\
{\small \textsuperscript{2}University of Carthage, Engineering School of Statistics and Information Analysis, Ariana, Tunisia,} \\
{\small \textsuperscript{3}Laboratory of BioInformatics bioMathematics, and bioStatistics (LR24IPT09), Institut Pasteur de Tunis,} \\ {\small University of Tunis El Manar, 13, place Pasteur, B.P. 74, Belvédère, 1002 Tunis, Tunisia} \\
\small{
   \textbf{Correspondence:} \href{mailto:hamza.landolsi@essai.ucar.tn}{hamza.landolsi@essai.ucar.tn}
  }
\\
}

\begin{document}
\vspace{1cm}
\maketitle
\vspace{3cm}

\begin{abstract}

The introduction of new features and services in the banking sector often overwhelms customers, creating an opportunity for banks to enhance user experience through financial chatbots powered by large language models (LLMs). We initiated an AI agent designed to provide customers with relevant information about banking services and insights from annual reports. We proposed a hybrid Customer Analysis Pipeline Retrieval-Augmented Generation (CAPRAG) that effectively addresses both relationship-based and contextual queries, thereby improving customer engagement in the digital banking landscape. To implement this, we developed a processing pipeline to refine text data, which we utilized in two main frameworks: Vector RAG and Graph RAG. This dual approach enables us to populate both vector and graph databases with processed data for efficient retrieval. The Cypher query component is employed to effectively query the graph database. When a user submits a query, it is first expanded by a query expansion module before being routed to construct a final query from the hybrid Knowledge Base (KB). This final query is then sent to an open-source LLM for response generation. Overall, our innovative, designed to international banks, serves bank's customers in an increasingly complex digital environment, enhancing clarity and accessibility of information.
\end{abstract}
\section{Introduction}

Banks all over the world are constantly introducing new features, delivering new financial services and providing their customer with clear booklets to guide them on how to use their mobile applications, how to benefit from all their available services, how to navigate their websites, etc. However when banks grow, even with the most straightforward user experience and with their efforts to make all informations accessible via promotions and accessible website, the task of naviagting and being informed with updates made the user lost in a maze \citep{ravalimanana2020impacts,shumba2023evaluation}. As the banking industry continues to embrace digital transformation, the potential applications of financial chatbots with Large Language Models (LLMs) are limitless presenting an exciting opportunity for banks to innovate and differentiate themselves in a rapidly changing market \citep{eisfeldt2023generative,ooi2023potential,accentureWaysGenerative}. \\

In fact, LLMs and GenAI have revolutionized the field of AI, allowing machines to create diverse content such as text \citep{doiGenerativeAIBased}, images \citep{springerImageGeneration}, music \citep{dong2024generative} and videos \citep{arxivSurveyGenerative}. Unlike discriminative models that classify, GenAI models generate new content by learning patterns and relationships from human-created datasets {\citep{arxivGenerativeProcess}}. \\ 

RAG framework is a solution to feed concise relevant context similar to the question and then passed to LLM to answer to a given query \citep{firstRAGpaper}. This come as a solution to overcome the context window of LLMs limitations. Recently a graph RAG-based approach \citep{procko2024graph} led to major improvements.  This method seeks to replicate the way our minds think and reason, similar to the functioning of a graph. When faced with a question that demands deep thought and reasoning, we begin to model the relationships among key factors, entities, and agents that may impact our area of interest. 
Many experts from the industry claim that agentic RAG is the future \citep{nguyen2024enhancing}: It aims to deliver a standalone solution that can be independent relying on multiple agents, specializing each in a defined workflow. 
With agentic style we can combine process and do multiple actions over a lot of sources and interact seamlessly with multiple agent or "expert" in their task. It leads to more context aware and more adaptability \citep{10.1145/3699824.3699832}. In this work, we designed an AI agent tasked to provide the customers with information related to the bank financial services, features, and key insights about the company through its annual reports. We obtained a hybrid approach using Vector and Graph RAG, to handle efficiently the nuances of both relationship based and contextual questions. 
Instead of thinking in a linear way that isolates each element, we connected various elements and asked questions that relate hidden factors. After iteratively refining our experiments and testing each component we conclude to our solution {Customer Analysis Pipeline RAG (CAPRAG)}. We discussed in next sections each workflow separately and the challenges and limitations of our solution. In Figure \ref{fig:customer-service-workflow} we gave the big picture of the workflow and we mentioned a Vector and Graph RAG pipelines. 


\section{Data source and description of bank ${\mathcal A}$}
\label{data}

The data used in this study is derived from publicly available documents and corporate publications pertaining to an international bank, that we anonymized as ${\mathcal A}$. Data sources include SEC filings, brochures and booklets.\footnote{SEC Filing are mandatory disclosures submitted by bank ${\mathcal A}$ to the U.S. Securities and Exchange Commission.} SEC filings contain comprehensive financial statements, management discussions, and risk factors, providing a detailed overview of the bank's financial performance and strategic outlook. From the SEC filings, we find annual reports containing qualitative data, such as bank ${\mathcal A}$’s strategic initiatives, corporate social responsibility (CSR) activities, and customer engagement programs. These qualitative insights offer a broader understanding of the bank's positioning and operational strategies beyond the financial statements. We extracted financial metrics such as total assets, net income, return on equity (ROE), and risk-weighted assets over a period of five years (2019–2023). These metrics are pivotal in evaluating the financial health of bank ${\mathcal A}$ and its performance relative to its peers. 
In addition to regulatory filings, bank ${\mathcal A}$ publishes various brochures and booklets that provide insights into its business operations, service offerings, and corporate governance. The brochures and booklets provided complementary information about bank ${\mathcal A}$ and that is very important for customers, providing every financial service offered by the bank, the criteria required, some frequent asked questions, and other relevant information. \\

Providing insights about these KPIs is essential for customers and the ones who are interested in investing in bank ${\mathcal A}$ as well. 
These materials, aimed at both clients and stakeholders, contribute to a qualitative data that complements the financial information. 

\begin{table*}[!ht]
\centering
\begin{tabularx}{\linewidth}{|p{4cm}|p{2.5cm}|X|}
\hline
{Data Source} & {Type of Data} & {Extracted Information} \\ \hline
SEC Filings & Quantitative & Financial metrics (ROE, assets, net income) / Strategic initiatives, CSR activities \\ \hline
Brochures and Booklets & Qualitative & Financial products/services, (loans, insurance, branches, etc.) \\ \hline
Annual Reports & Quantitative\newline Qualitative & Financial performance, governance insights \\ \hline
Sustainability Reports & Qualitative & ESG goals, sustainability programs \\ \hline
\end{tabularx}
\caption{Summary of data sources and extracted information}
\label{tab:data_summary}
\end{table*}

Each of these sources provides unique insights into bank ${\mathcal A}$'s operations and strategy, and together they form a comprehensive dataset for analysis in this study (Table \ref{tab:data_summary}). 


\section{Evaluation pipeline: LLM as a judge} \label{evaluation}

There are several ways to monitor and track the evolution of our work, for this project we used LLMs to help us rate and assess our metrics of interest. This technique is called "LLM as a judge". \\

We setup an evaluation pipeline in an automatic way. First of all, we choose the metrics that align the best with our challenges. To ensure the effectiveness and reliability of the RAG model, it is essential to evaluate its outputs using specific metrics. The three metrics we focus on are answer relevance, context relevance and groundedness \cite{es2024ragas,roychowdhury2024evaluation,yu2024defense}.

\begin{definition}[Answer Relevance]
    Answer relevance measures how accurately the generated response addresses the given query. A high answer relevance score indicates that the response is directly related to the question and provides useful information.
\end{definition}

\begin{definition}[Context Relevance]
    Context relevance evaluates the extent to which the generated response aligns with the context provided. This metric ensures that the response is coherent within the given context and maintains the flow of the conversation or narrative.
\end{definition}

\begin{definition}[Groundedness]
    Groundedness assesses whether the generated response is based on factual information retrieved from reliable sources. A grounded response is one that can be traced back to authoritative references, ensuring the accuracy and trustworthiness of the information.
\end{definition}

\begin{figure}[!ht]
    \centering
    \includegraphics[width=\linewidth]{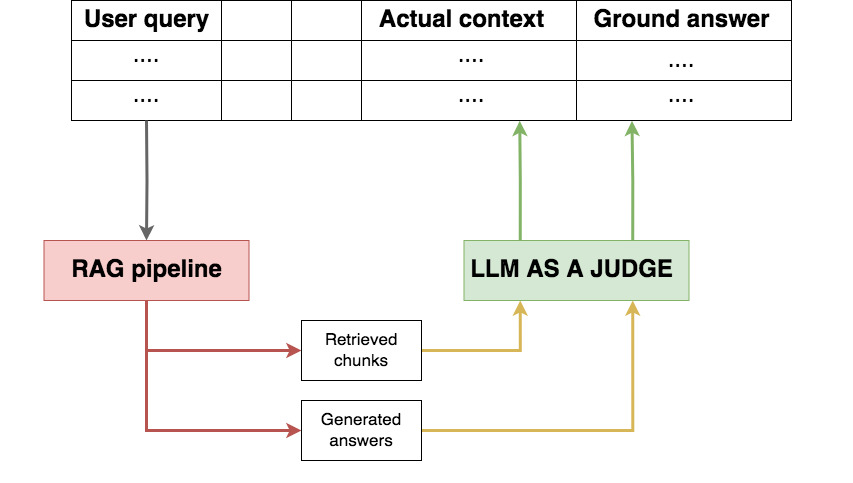}
    \caption{Evalutation pipeline with LLM as judge}
    \label{fig:eval_llm_judge}
\end{figure}

The evaluation pipeline plays an important role in the development process. By systematically assessing the model's outputs based on answer relevance, context relevance, and groundedness, we can ensure that the model generates high-quality, accurate, and coherent responses (see Figure \ref{fig:eval_llm_judge}). We improve the user experience and builds trust in the system's capabilities. 


\section{Vector RAG: Methodology and Results}

\begin{figure}[!ht]
    \centering
    \includegraphics[width=0.59\linewidth]{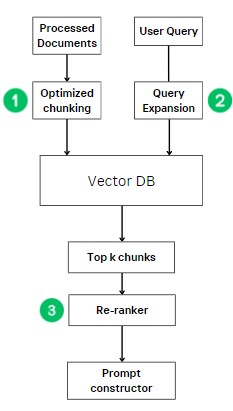}
    \caption{Vector RAG enhancements}
    \label{fig:vector-rag}
\end{figure}

As our main focus is improving customer service efficiency in the banking industry, we leveraged advanced chunking optimization techniques, query expansion, and retrieval enhancement strategies \cite{finardi2024chronicles,modi2024automated}. The goal is to improve the accuracy and speed of responses to customer inquiries by optimizing how Knowledge Base (KB) articles are retrieved and processed. In this Section we detailed further the Vector RAG component (See Figure \ref{fig:vector-rag}). 

We first explored various chunking optimization techniques to segment large KB into smaller, more relevant pieces for retrieval (we have chosen the semantic chunking for our solution).  
Second, we utilized query expansion strategy to refine customer queries, making them more robust and comprehensive for retrieval systems. These techniques allow for a better match between the customer query and relevant chunks. 
For the retrieval system, we compare different retrieval enhancement techniques tailored to the banking domain. We choose finally the FlashRank re-ranker \citep{githubGitHubPrithivirajDamodaranFlashRank} as a post retrieval enhancement strategy. \\

We used Zephyr for generation in our Vector RAG mechanism.\footnote{\citep{doiZephyrDirect} we aim to achieve comparable results with other big LLMs which offer enterprise-ready solutions in terms of scalability, security, and license compliance, especially relevant in the banking sector.} \\

The results in Table \ref{fig:rag_eval} compares the performance of various techniques used on manually created dataset to optimize the Retrieval-Augmented Generation (RAG) model across the three evaluation metrics: {Answer relevance}, {Context relevance}, and {Groundedness}.

\begin{table*}[!ht]
    \centering
    \begin{tabular}{|l|c|c|c|}\hline 
         & Answer Relevance & Context Relevance & Groundedness  \\ \hline
         Baseline RAG & && \\ \hline 
         + Chunking Optimization &&& \\ \hline 
         + Enhancing Retreival &&& \\ \hline 
         + Query Translations &&& \\ \hline 
    \end{tabular}
    \caption{Comparison of Techniques on Different Metrics for Vector RAG}
    \label{fig:rag_eval}
\end{table*}
\begin{figure}[!ht]
    \centering
    \includegraphics[width=0.9\linewidth]{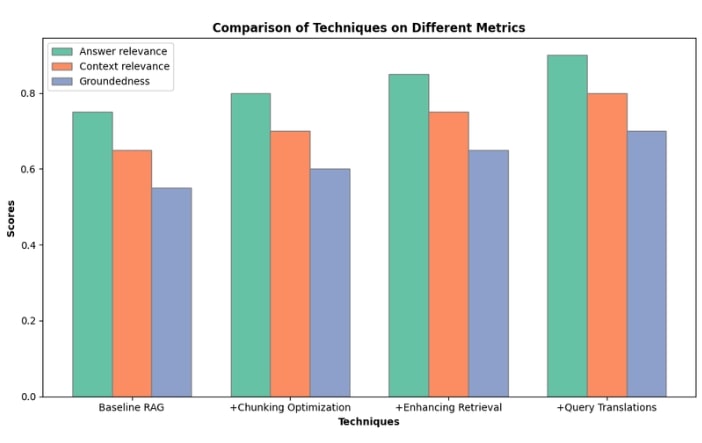}
    \caption{Vector RAG evalutaion}
\end{figure}

\paragraph{Baseline RAG.} As expected, the baseline configuration shows moderate scores across all metrics. While it performs fairly well in terms of Answer relevance (about 0.75), the Groundedness score is significantly lower (around 0.55), which indicates that the model has room for improvement when it comes to grounding answers in retrieved documents.  
\paragraph{Chunking Optimization.} Introducing chunking optimization leads to an increase in both Answer relevance and Context relevance. The {Answer relevance} improves slightly compared to the baseline, suggesting better handling of long documents by splitting them into smaller chunks. However, the Groundedness score slightly decreases, implying that chunking may inadvertently lead to less reliance on the retrieved documents for the final answer.

\paragraph{Enhancing Retrieval.} This optimization strategy yields the best overall Answer relevance score, exceeding 0.8, indicating that improved retrieval techniques help the model generate more accurate answers. However, similar to chunking, the Groundedness remains relatively lower, indicating that although retrieval improves answer quality, it may not always improve the model's alignment with retrieved evidence.

\paragraph{Query Translations.} This technique leads to a balanced improvement across all metrics. Answer relevance remains high, and Context relevance sees a boost, indicating that better query formulation helps retrieve more relevant context. Additionally, Groundedness improves compared to the baseline, suggesting that query translations contribute to generating more factually grounded answers. \\

Overall, the results indicate that while optimizations like chunking and enhanced retrieval can improve answer quality, further work may be needed to improve the {Groundedness} of the model's responses, ensuring the generated answers align more closely with the retrieved documents. Techniques like query translations show promise in improving performance across the board.

\section{Graph RAG}

The first step in creating a powerful graph knowledge base is to model the data in the most appropriate way (to capture the complex relationships between chunks).
The graph should be created with nodes (representing the entity) and edges (the relationship between two nodes). 

\subsection{Schema Construction}
\label{schema construction}

We first created nodes of label {Document}, where in each case we could store the document name and other important metadata as properties. The second type of node is Section, where we store sections of each document. Since documents can have sections and many hierarchical subsections, we can handle this by adding two relationships:
\begin{enumerate}
    \item Section to Section relationship: \textcolor{blue}{\tt UNDER-SECTION} that link the subsection to its parent section;
    \item Section to Document relationship: \textcolor{blue}{\tt HAS-DOCUMENT} which can link sections to the documents they are located in.
\end{enumerate}

\begin{figure}[!ht]
\centering
\includegraphics[width=\textwidth]{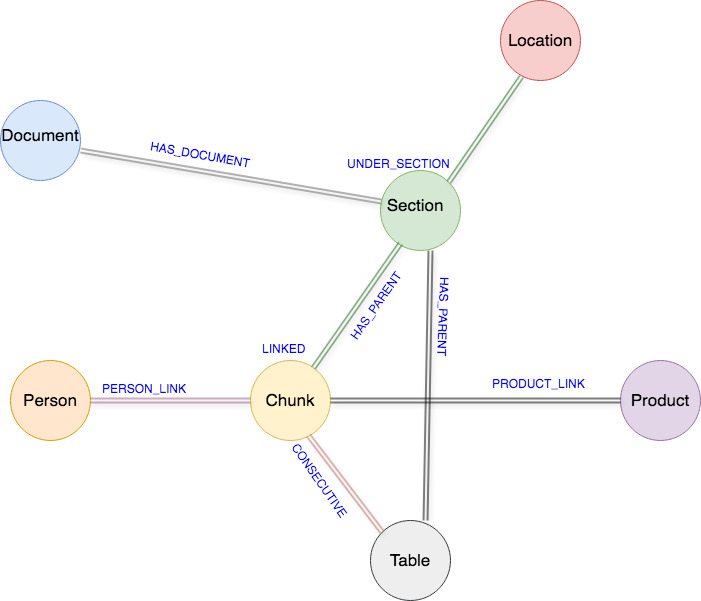} 
\caption{Expanded graph schema using Graph Knowledge}
\label{fig:graph2}
\end{figure}

This gives a comprehensive overview for our hierarchy of the corpus layout to our knowledge base. 
The third type of node is Chunk. Here we made special node for Table chunks, so we preserve only table related information and separate them from other information (without loosing the scope and the context around the table). We can do so by establishing new edges: \textcolor{blue}{\tt HAS\_PARENT}, \textcolor{blue}{\tt CONSECUTIVE}, \textcolor{blue}{\tt PERSON\_LINK}, \textcolor{blue}{\tt PRODUCT\_LINK}, etc. For example, the relationship between Table and Section is given by \textcolor{blue}{\tt HAS\_PARENT}; \emph{idem} for Chunk to Section relationship. The Chunk to Chunk relationship is given by the edge  \textcolor{blue}{\tt LINKED} (that link each chunk with the next chunk in corpus). This can be a solution for the problem of lengthy chunks that lead to loss of information. When we divide them to two chunks carrying the information of the same product or service (information loss or chunk corruption). The Table to Chunk relationship is given by \textcolor{blue}{\tt CONSECUTIVE}, which links Tables to their next and previous Chunks or other Tables.



Let focus on some of the powerful entities deemed insightful for our use case. Figure \ref{fig:graph2} visualizes our expanded graph at this stage. Person is often a very important entity in our corpus that we should focus on. With the length of our chunks, this crucial information may be diluted in the embedding process hence we lose this information. 
For the Location entity, we focused on the geospatial coordinate (Longitude and Latitude) rather than only working with its semantic vector representation. We made the LLM responsible for text-to-cypher mission the ability to construct Cypher queries capable of solving spatial metric related questions. \\

Creating a knowledge graph - ready to integrate with our RAG framework - passed through modeling our data base schema and populating our graph database. 

\subsection{Methodology}

Although semantic similarity is very powerful, it might fail also to identify other types of similarity such as {geospatial similarity}. For example, the user query might be: \texttt{"I am in Ariana and I am wondering what's the nearest bank ${\mathcal A}$.\footnote{We took bank ${\mathcal A}$ as our fictive bank customer} ATM branch to me?"}.

The complexity of the graph database come with more burden and challenges. Having vector representation of embedded user question doesn't help us to query our knowledge graph. Thus,
we apaplied LLMs for generating Cypher queries from the user query.\footnote{Cypher is the language used to query a graph database, such as Neo4j \citep{neo4j2017}.}  
Achieving this task can be through choosing a powerful LLM, expanding its context window with tailored prompting. The option to go for this use case is often one of the top performing LLMs in the leaderboard.

\begin{table}[!ht]\begin{scriptsize}\begin{center}
\begin{tabular}{|p{2cm}|p{4.5cm}|p{8cm}|}
  \hline
   User query & Vector KB (Knowledge Base) based RAG & Graph KB RAG \\
  \hline
  {\tt Can you explain the professional background and career progression of the key executives mentioned in the bank ${\mathcal A}$ annual reports? Do they have any common traits or experiences that have shaped their leadership at bank ${\mathcal A}$?} & {\tt Based on context provided, the key executives include:
  Maria R. - Chief Executive Officer
  Jason Q. - Group Financial Director
  Peter M. - Deputy CEO
  They have played significant roles in the company's strategic decisions and overall performance. For more detailed information, please refer to the bank ${\mathcal A}$ website or contact the team. There is no specific information available about common traits or experiences.} & {\tt From the bank ${\mathcal A}$ annual reports, we have identified the following details regarding the professional background and career progression of key executives, as well as their common traits:
  Maria R. - Chief Executive Officer
  Professional Background: Maria R. has a distinguished career in both the public and private sectors. She previously served as the Director-General of South Africa's National Treasury.
  Career Progression: Maria joined bank ${\mathcal A}$ in 2009 and has since led several key initiatives, including the transformation of the bank's digital strategy and expansion into new markets.
  Jason Q. - Group Financial Director
  Professional Background: Jason Q. is a qualified Chartered Accountant with over 20 years of experience in the financial sector. Before joining bank ${\mathcal A}$, he held senior positions at Deloitte.
  Career Progression: Jason has been with bank ${\mathcal A}$ since 2008, initially as the Head of Finance for Retail and Business Banking before being appointed Group Financial Director in 2016.
  Peter M. - Deputy CEO
  Professional Background: Peter M. has an extensive background in both corporate leadership and media. Prior to bank ${\mathcal A}$, he was the CEO of Tiger Brands and the SABC.
  Career Progression: Joining bank ${\mathcal A}$ in 2016, Peter has been pivotal in driving the bank's growth across the African continent.
  Common Traits and Experiences
  Public and Private Sector Experience: Both Maria R. and Peter M. have extensive experience in both public and private sectors, which has contributed to their versatile leadership styles.
  Leadership in Transformation and Expansion: All three executives have played significant roles in transformative projects and expansion initiatives within bank ${\mathcal A}$, showcasing their strategic leadership and vision.} \\
  \hline
\end{tabular}    \end{center}
\end{scriptsize}
\caption{Comparison between Vector and Knowledge RAG output}
\label{tab:prompt}
\end{table}

Many prompting techniques have emerged to make the best of their capabilities \citep{wei2022chain,zhang2022auto,promptingguide2022,tian2024graph}. One considerable limitation of this is that LLM try to follow our instructions and give us relevant and most of the time very accurate results. But when it comes to specifying a valid rigid format, it often fall to respect it. \\
For example, in our use case, we feed LLM user query and graph knowledge database schema as input with a prompt template guiding it to output a valid Cypher query in string format. 
As we didn't have enough resources to fine tune our own custom model or use some specialized closed source models so we designed a Cypher repository, each Cypher template in this repository with a description so that we can decide if the user query can be answered via this Cypher or not.

\section{Customer Analysis Pipeline RAG (CAPRAG)} 
\label{caprag}

\begin{figure}[!ht]
    \centering
    \includegraphics[width=0.7\linewidth]{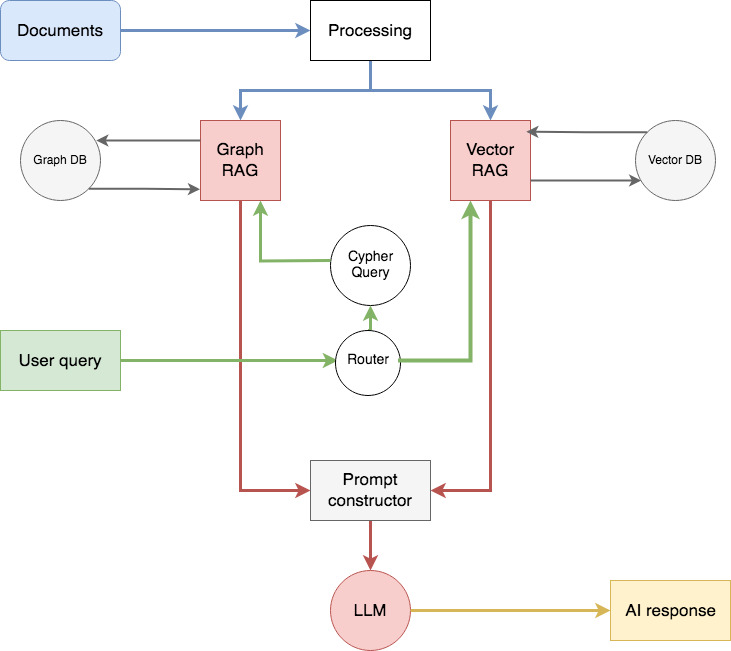}
    \caption{Workflow of CAPRAG}
    \label{fig:customer-service-workflow}
\end{figure}

We aim to setup a processing pipeline that refine the text data. Than we simultaneously used this processed data in two main pipelines {Vector RAG} and {Graph RAG}, see Figure \ref{fig:customer-service-workflow}. Finally we populated respectively the vector database and graph database with the final ready data to use for retrieval.

The router is a LLM prompted to divide the query or reformulate it to each of the next components the graph RAG and the Vector RAG.

The Cypher query component is responsible for querying graph database, and we tried several techniques to handle text-to-cypher problem \citep{francis2018cypher}.\footnote{We utilized a static approach due to hardware requirements and private source LLM provider API keys restrictions.}
So when a user enter his question, we pass the query to a Query Expansion module that will enrich our user query and send it to a router where he constructs a query from the hybrid knowledge base and finally send the final prompt to the open source LLM for generation. \\

We used a decoder only based LLM architecture \citep{brown2020language}. We employed {\tt HuggingFace H4/zephyr-7b-beta} model checkpoint, configured to generate text \citep{jain2022hugging}, via {\tt HuggingFaceEndpoint}. 
It is configured with a maximum length of 2000 tokens and a temperature of 0.01, which means we are encouraging the LLM to be more deterministic and not allowing it to be creative.\footnote{This is a hyper-parameter for adjusting the creativity of generation, i.e. pushing the model to be greedy when it comes to choose the highest probability of next predicted token.} 
In our use case, the LLM sticks to the provided context and reduces the hallucination as much as possible, see Table \ref{tab:prompt}. \\

\section{Discussion}

In this study, we highlighted the limitations of using embedded user queries for retrieving chunks based solely on cosine similarity with their vector representations. This approach is inadequate for capturing complex relationships and geo-spatial information. Consequently, alternative methods may be necessary to enhance the effectiveness of information retrieval in these contexts. 
Our findings indicated that, although several automatic expansion techniques are available, manual expansion can be optimal and preferable. This is particularly true for specific industries. 


While many researchers have utilized graph RAG for ontology-based graph construction in the banking sector, there are notable variations in their approaches. For instance, \cite{KGbankincident} explored novel methods for extracting accurate and insightful entities. Additionally, some studies have focused on establishing the relationships between sections and the structure of the document hierarchy. These diverse methodologies contribute to a deeper understanding of how to effectively leverage graph RAG in this domain. Other research efforts have concentrated on geospatial data, achieving significant advancements in enhancing embeddings that capture both semantic and spatial meanings \cite{hu2024geometric}. Some studies have focused on customizing agentic workflows to facilitate interactions with geographic data \cite{agenticGIS}. These contributions highlight the importance of integrating semantic and spatial dimensions in geospatial analysis.\\

Our findings complement the work of \cite{hybridRAG}, which aims to enhance questions related to spatial terms. This enhancement requires metrics of similarity that extend beyond semantic-oriented measures, such as cosine similarity. For that, we are letting a router LLM to think what metric to choose and which Cypher query to run. We pave the way for innovative approaches that rely on alternative similarity metrics. Our method begins by identifying issues in similarity that cannot be captured semantically. We then extract the relevant entities associated with these problems. Finally let the LLM decide how to query these nodes of interest. 
The extraction of product entities can give banks and financial institutions the opportunity of either up-selling or cross-selling their products based on the customer profile. As ontology based graph knowledge is gaining a huge success proven by recent research works, there is a growing need for identifying relevant entities for each sector. 
Relying on curated factors and entities from domain experts can be more effective than allowing LLMs to construct the knowledge graph independently. An ideal approach may involve combining both methods. We have developed a pipeline that intelligently selects the appropriate path and components for answering specific questions. This pipeline includes a router that understands the requirements of each question. This ensures a more accurate and relevant response. \\

As we don't have access to closed source models, and the hardware required to run open source LLMs, we focused on using open source only models. We were limited by the Hugging Face maximum quota of requests. So we couldn't evaluate our pipelines on open data benchmarks in this field or conclude to comparisons. We could potentially get better results in extracting entities of our focus by experimenting with specialized LLMs or agents for converting human-like queries to Cypher given our proposed scheme. It may be possible to use a structured output format more easily, using features that are only supported in closed source models.

\section{Conclusion}
Our research highlights the integration of Vector RAG and Graph RAG in an advanced pipeline, for an application into the banking sector. By refining text data and utilizing a hybrid approach, we have developed a robust Customer Analysis Pipeline RAG (CAPRAG) that effectively addresses the complexities of customer inquiries. This innovative solution not only enhances the retrieval of relevant information but also fosters deeper connections between key factors and entities within the banking ecosystem. Our AI agent serves as a valuable tool for providing timely insights into financial services, thereby empowering customers to make informed decisions. 




\section*{Acknowledgments}

We would like to thank the anonymous reviewers for their valuable comments and encouragement in improving this work.






\end{document}